\documentclass[runningheads]{llncs}
\usepackage{graphicx}

\usepackage{subcaption}
\usepackage{amsmath}
\makeatletter
\newcommand*{\rom}[1]{\expandafter\@slowromancap\romannumeral #1@}
\usepackage{amsfonts}

\begin{document}

\title{Multimodal Sensor Fusion In Single Thermal image Super-Resolution
\thanks{This work was supported by the European Regional Development Fund (ERDF) and the Brussels-Capital Region within the framework of the Operational Programme 2014-2020 through the ERDF-2020 project F11-08 ICITY-RDI.BRU.
The Titan X Pascal used for this research was donated by the NVIDIA Corporation. We are grateful to Thermal Focus BVBA for their help and support.}
} 
\titlerunning{Multimodal Fusion In Thermal SR} 


\author{Feras Almasri\inst{1} 
\and
Olivier Debeir\inst{2}
}
%

\authorrunning{F. Almasri et al.} 


\institute{Dept.LISA - Laboratory of Image Synthesis and Analysis, \\
Université Libre de Bruxelles\\
CPI 165/57, Avenue Franklin Roosevelt 50, 1050 Brussels, Belgium.
\email{falmasri@ulb.ac.be}, \email{odebeir@ulb.ac.be}
}

\maketitle

\begin{abstract}
With the fast growth in the visual surveillance and security sectors, thermal infrared images have become increasingly necessary in a large variety of industrial applications. This is true even though IR sensors are still more expensive than their RGB counterpart having the same resolution. In this paper, we propose a deep learning solution to enhance the thermal image resolution. The following results are given: (\rom{1})  Introduction of a multimodal, visual-thermal fusion model that addresses thermal image super-resolution, via integrating high-frequency information from the visual image. (\rom{2}) Investigation of different network architecture schemes in the literature, their up-sampling methods, learning procedures, and their optimization functions by showing their beneficial contribution to the super-resolution problem. (\rom{3}) A benchmark ULB17-VT dataset that contains thermal images and their visual images counterpart is presented. (\rom{4})  Presentation of a qualitative evaluation of a large test set with 58 samples and 22 raters which shows that our proposed model performs better against state-of-the-arts.

\end{abstract}
\section{Introduction}
In digital images, what we perceive as details greatly depends on the image resolution. The higher the resolution the more accurate the measurement. The visible RGB image has rich information, but objects can occur in different conditions of illumination, occlusion and background clutter. These conditions can severely degrade the system's performance. Therefore visible data is found to be insufficient and thermal images have become a common tool to overcome these problems. Thermal images are used in industrial processes such as heat and gas detecting and they are also used to solve problems such as object detection and the self-driving car.

Integrating captured information from different sensors such as RGB and thermal offers rich information to improve the system performance. In particular, when the nature of the problem requires this integration, and when the environmental conditions are not optimal for a one sensor approach, multimodal sensor fusion methods have been proposed ~\cite{cho2014multi,qu2017active}. However, thermal sensor cost grows significantly with the increase of its resolution and it is primarily used in low-resolution and in low contrast which introduces the necessity to obtain a higher resolution sensor ~\cite{choi2016thermal}. As a result, a variety of techniques in computer vision have been developed to enhance thermal resolution given their low-resolution counterpart.

The single super-resolution problem has been widely studied and well-defined in computer vision ~\cite{chen2016color,panagiotopoulou2008super,kiran2017single}. It is defined as non-linear mapping and prediction of a high-resolution image (HR) given only one low-resolution image (LR). However, this is an ill-posed problem since it is a one-to-many problem. Given that multiple HR images it is possible to produce a single LR image, thus mapping from LR to SR is to recover the lost information giving only the information in the LR image. Though they achieved high performance, these methods are limited by their handcrafted features techniques.

Recently with the development of the convolutional neural network (ConvNet), several methods have shown the ability to learn high non-linear transformations. Rather than using handcrafted features, the ConvNet model is capable of automatically learning rich features from the problem domain and adapt its parameters by minimizing the loss function. Most recently,the ConvNet model has been widely used in the SR problem and achieved new high performance. Despite significant progress, the proposed solutions still suffer from the lack of ability to recover high-frequency information.

Most SR conventional methods focus on measuring the similarity between the SR image and its ground truth via pixel-wise distance measurement, although the reconstructed images are blurry by missing sharper edges and texture details. However, this problem is the fault of the objective function, as the classical way to optimize the target is to minimize the content loss by minimizing the mean square error (MSE) loss function. By definition this finds the average values in the HR manifold and consequently maximizes the Peak signal-to-noise ratio (PSNR). By only applying the content loss function, the low-frequency information is restored, but not the high-frequency information. However, MSE is limited in preserving the human visual perception, and PSNR measurement cannot indicate the SR visual perception ~\cite{kiran2017single,2016arXiv160904802L}.

Different approaches such as perceptual loss ~\cite{johnson2016perceptual} and adversarial loss ~\cite{ledig2016photo},have been proposed to address this drawback, and have shown important progress in recovering high frequency details. Instead of only doing a pixel-wise distance measurement, a mixture of these loss functions could generate high-quality SR images. Also, different model schemes have shown higher image quality such as learning the residual information ~\cite{kim2016accurate} or by gradually up-sampling ~\cite{lai2017deep}.

The primary focus of this work was to build a deep learning model that applies multimodal sensor fusion using visible (RGB) and thermal images. The model should integrate the two inputs and enhance the thermal image resolution. The latter part is inspired by the recent advances in RGB super-resolution problem.A thermal GAN based framework is proposed, to enhance the LR thermal image by integrating the rich information in the HR visual image. However, the HR visual sensor price is considerably low compared to the LR thermal sensor and it captures extra information taken from a different domain. We show that HR visual images can help the model fill the missed values and generate higher frequency details in the reconstructed SR thermal image. The model proposed uses the content loss to preserve the low-frequency image content, and the adversarial loss to integrate the high-frequency information.


\section{Related Works}
\begin{figure}
    \centering
    \begin{subfigure}[b]{0.25\textwidth}
        \includegraphics[width=\textwidth]{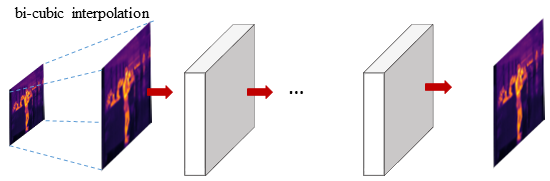}
        \caption{SRCNN \cite{2015arXiv150100092D}}
    \end{subfigure}\quad
    \begin{subfigure}[b]{0.25\textwidth}
        \includegraphics[width=\textwidth]{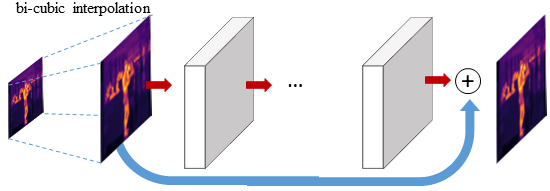}
        \caption{VDSR \cite{kim2016accurate}}
    \end{subfigure}\quad
    \begin{subfigure}[b]{0.25\textwidth}
        \includegraphics[width=\textwidth]{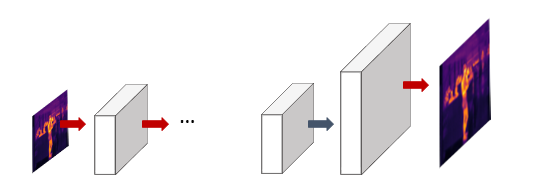}
        \caption{FSRCNN \cite{2016arXiv161109969Y}}
    \end{subfigure}\\
    
    \begin{subfigure}[b]{0.25\textwidth}
        \includegraphics[width=\textwidth]{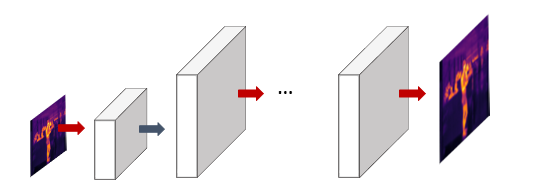}
        \caption{SRCNN-Dconv}
    \end{subfigure}\quad
    \begin{subfigure}[b]{0.25\textwidth}
        \includegraphics[width=\textwidth]{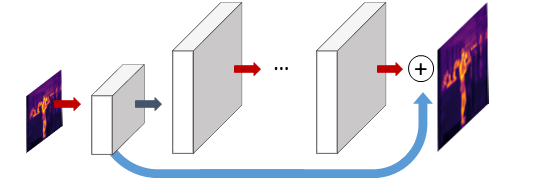}
        \caption{VDSR-Deconv}
    \end{subfigure}\quad
    \begin{subfigure}[b]{0.25\textwidth}
        \includegraphics[width=\textwidth]{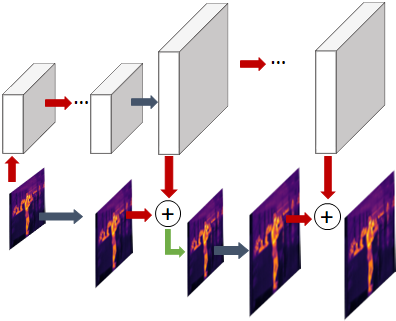}
         \caption{LAPSRN \cite{lai2017deep}}
    \end{subfigure}\quad
    
    \caption{Network architecture schemes.}\label{fig:modelAbstract}
\end{figure}

Thus far, a number of studies in computer vision and machine learning have been proposed. This discussion focuses on example-based ConvNet methods. Fig.~\ref{fig:modelAbstract} depicts the different model architecture schemes, their up-sampling method, and their learning procedure.

\textbf{A. Resolution Training}. The model can be trained to extract features from an up-sampled image in direct mapping using these features to produce SR image as in ~\cite{2015arXiv150100092D}. The input is either pre-processed, using an interpolation method as shown in Fig \ref{fig:modelAbstract} (a), or up-sampled using trainable parameters as shown in Fig \ref{fig:modelAbstract} (d). In another approach, the model can extract features directly from the low-resolution image and map them into high-resolution by using up-sampling techniques at the end of the network as in ~\cite{2016arXiv161109969Y}, this model, shown in Fig \ref{fig:modelAbstract} (c) accelerates the model performance.

\textbf{B. Residual Learning and Supervision}. Since super-resolution output is similar to the low-resolution input,with the high-frequency information missing, the learning can be made to produce only the residual information. VDSR \cite{kim2016accurate} and DRSN ~\cite{kim2016deeply} trained a model that learns residual information between LR and HR images. They used a skip connection, that adds the input image to the model residual output, to produce SR image. Lai et al ~\cite{lai2017deep} found that reconstructing the SR image immediately with a high up-sampling scale is a challenging problem. Therefore they addressed the problem in a gradual up-sampling procedure, using deep supervision in each up-sampling scale, and a residual learning as shown in Fig. \ref{fig:modelAbstract} (f).

\textbf{C. Up-sampling Procedure}. A mixture of network architectures and learning procedures can be used with different up-sampling methods. In SRCNN ~\cite{2015arXiv150100092D} and VDSR ~\cite{kim2016accurate}, the network takes an interpolated image as input, using either a bilinear or bicubic interpolation, or a Gaussian-Spline kernel ~\cite{huang2018densely}. The up-sampling can be learnable and accelerated by using a transposed convolution (deconvolution) layer as in Fig \ref{fig:modelAbstract} (d) and (e) which is a modified version of SRCNN and VDSR, or added to the end of the network as in FSRCNN Fig \ref{fig:modelAbstract} (c).  A trainable Sub-pixel convolution (Pixelshuffle) ESPCN ~\cite{shi2016real} is also used at the end of the network to up-sample input features as in ~\cite{ledig2016photo}.

\textbf{D. Optimization Function}. 

The optimization procedure seeks to minimize the distance between the original HR image and the generated SR image. The most used optimization function in the SR problem is the content loss, which is done using the MSE as in ~\cite{kim2016accurate} or Charbonnier as in ~\cite{lai2017deep}. SRGAN \cite{2016arXiv160904802L} instead uses the adversarial loss and ~\cite{johnson2016perceptual} uses the perceptual similarity loss to enhance the reconstructed image resolution.


\section{Proposed Methods}
In this work, the first ConvNet that integrates visual-thermal images to generate thermal image super-resolution is described. Our main contributions are:
\begin{itemize}
    \item[$\bullet$] Unlike the RGB-based super-resolution problem, advancement in thermal super-resolution is still relatively low. Therefore, few benchmarks in thermal image SR  ~\cite{panagiotopoulou2008super} and they are rarely available. To this end, the authors created a benchmark ULB17-VT multimodal sensors dataset that contains thermal images and their visual images counterparts.
    \item[$\bullet$] The model is inspired by the work in ~\cite{2016arXiv160904802L}. Different modified network architecture schemes, up-sampling methods and optimizing functions are investigated, to verify our model contribution with reference to the current improvement in the super-resolution problem literature.
    \item[$\bullet$] Confirmation is given which shows that this thermal SR model, which integrates the visual-thermal fusion, does generate higher human perceptual quality images. This improvement is due to the rich details in the visual images, and the common relation with their thermal image counterpart.
    \item[$\bullet$] A qualitative evaluation method based on 58 samples, which is a large qualitative evaluation in the SR problem domain, was also used to test how well the model performed. Twenty-two people were asked to rate the better generated image. The results of study shows that the proposed model performs better against state-of-the-art methods.
\end{itemize}

\subsection{ULB17 Thermal Dataset}
A FLIR-E60 camera with multimodal sensors (thermal and color) was used. The thermal sensor produces (320 x 240) pixel resolution with $0.05^{\circ}$C thermal sensitivity and $-20^{\circ}$C to $650^{\circ}$C which provides good quality thermal images. The color sensor is 3 megapixels and produces (2048 x 1536) pixels resolution. This device allows  capture of both thermal and RGB images aligned with the same geometric information simultaneously. Thermal images were extracted in their raw format and logged in 16-bit float per-pixel in one channel in their raw format, in contrast with ~\cite{hwang2015multispectral} in which samples are post-processed and compressed into an uint8 format. All samples in this study are aligned automatically by the device supported software with their associated cropped color images of size (240 x 320) pixel with the same geometry.

\begin{figure}
    \centering
    \begin{subfigure}[b]{0.24\textwidth}
        \includegraphics[width=\textwidth]{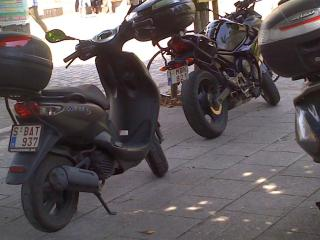}
    \end{subfigure}
    \begin{subfigure}[b]{0.24\textwidth}
        \includegraphics[width=\textwidth]{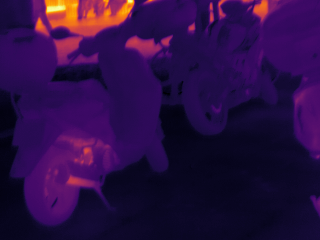}
    \end{subfigure}
    \begin{subfigure}[b]{0.24\textwidth}
        \includegraphics[width=\textwidth]{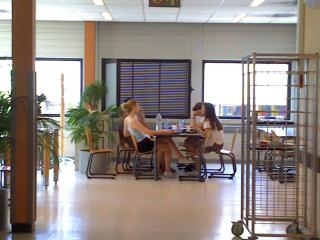}
    \end{subfigure}
    \begin{subfigure}[b]{0.24\textwidth}
        \includegraphics[width=\textwidth]{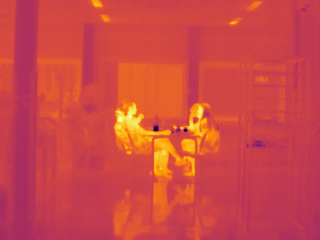}
    \end{subfigure}\\
    
    \begin{subfigure}[b]{0.24\textwidth}
        \includegraphics[width=\textwidth]{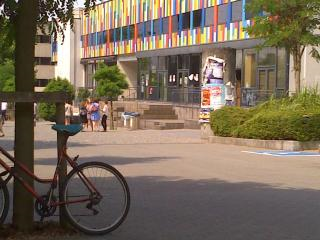}
    \end{subfigure}
    \begin{subfigure}[b]{0.24\textwidth}
        \includegraphics[width=\textwidth]{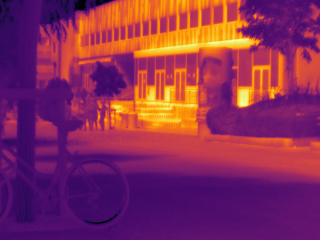}
    \end{subfigure}
    \begin{subfigure}[b]{0.24\textwidth}
        \includegraphics[width=\textwidth]{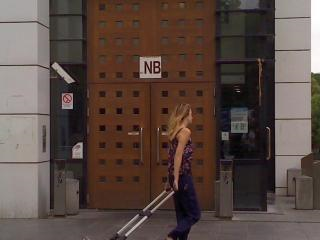}
    \end{subfigure}
    \begin{subfigure}[b]{0.24\textwidth}
        \includegraphics[width=\textwidth]{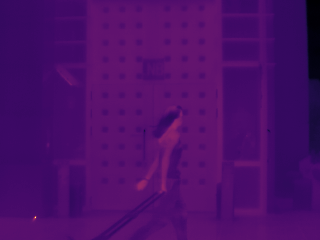}
    \end{subfigure}\\
    
    \begin{subfigure}[b]{0.24\textwidth}
        \includegraphics[width=\textwidth]{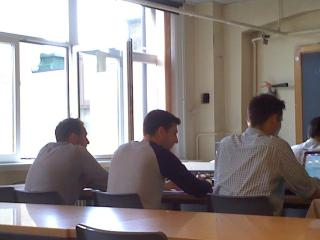}
    \end{subfigure}
    \begin{subfigure}[b]{0.24\textwidth}
        \includegraphics[width=\textwidth]{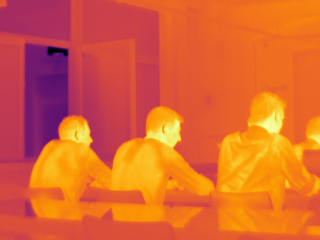}
    \end{subfigure}
    \begin{subfigure}[b]{0.24\textwidth}
        \includegraphics[width=\textwidth]{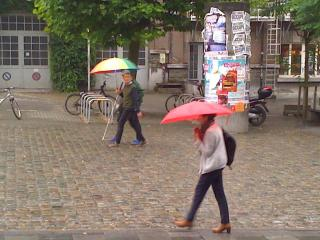}
    \end{subfigure}
    \begin{subfigure}[b]{0.24\textwidth}
        \includegraphics[width=\textwidth]{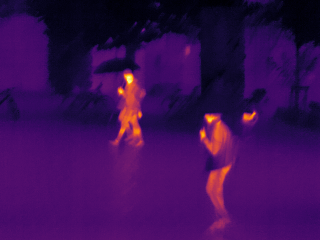}
    \end{subfigure}
    \caption{Visual-Thermal samples form our ULB17-VT benchmark.}\label{fig:dssamples}
\end{figure}

Images in our benchmark were captured inside ULB\footnote{\label{footnote:ULB} Université Libre de Bruxelles} campus. Each image acquisition process took approximately 3 seconds, which made the data acquisition process rather slow. The acquisition was made in different scenes and different environments (indoor and outdoor, during winter and summer and with static and moving objects) as shown in Fig.~\ref{fig:dssamples}. Thermal and RGB images were manually extracted and annotated with a total of 570 pairs. The framework is divided into 512 training and validation samples and 58 test samples.

\subsection{Proposed framework}
In this section our model methodology including how different model schemes, up-sampling methods, and optimization function are essential in producing better human perceptual SR images are described.

The aim is to estimate the HR thermal image $T^{HR}$ from its LR counterpart $T^{LR}$ by a 4x factor. The $T^{LR}$ images are produced by first applying Gaussian pyramids to $T^{HR}$ samples and they are then down-scaled by a factor of .25x, from (240 x 320) pixels to (60 x 80) pixels.

Our proposed model belongs to the model scheme of FSRCNN shown in Fig \ref{fig:modelAbstract} (c). The core of our model scheme is to perform feature extraction and feature mapping on the image original size. The model is constructed using X residual blocks with an identical layout inspired by SRGAN ~\cite{2016arXiv160904802L}. The model then up-samples the feature maps with two sub-pixel convolution layers as proposed by ~\cite{shi2016real}. Starting from the main model as a baseline,the model gain was investigated as follows:

\begin{itemize}
    \item[$\bullet$] Instead of up-sampling the features at the end of the network, we tested the model scheme of SRCNN \cite{2015arXiv150100092D} shown in Fig \ref{fig:modelAbstract} (d). Due to investigation of high-resolution training methods, Sub-pixel layers are removed from this model, and two Deconv layers are added at the beginning of the network by a factor of 2.
    \item[$\bullet$] The residual scheme proposed by VDSR as shown in Fig \ref{fig:modelAbstract} (b) and (e) is tested using (\rom{1}) bi-linear interpolation, (\rom{2}) bi-cubic interpolation, and (\rom{3}) different trainable up-sampling layers.
    \item[$\bullet$] Visual images are integrated to investigate whether their texture details can enhance the thermal SR generating process. 
    \item[$\bullet$] Recently, the Generative adversarial network (GAN) \cite{goodfellow2014generative} has witnessed creative implementation in several tasks \cite{isola2017image,chen2017face,zhang2017stackgan,zhu2017unpaired,wu2017srpgan}. GAN provides a powerful framework which consists of an image transformer and an image discriminator, that can generate images with high human perceptual quality and are similar to the real images distribution. To achieve GAN contribution, our baseline model is re-trained on GAN based model.
\end{itemize}

\begin{figure}
    \centering
    \begin{subfigure}[b]{\textwidth}
    \centering
        \includegraphics[width=.9\textwidth]{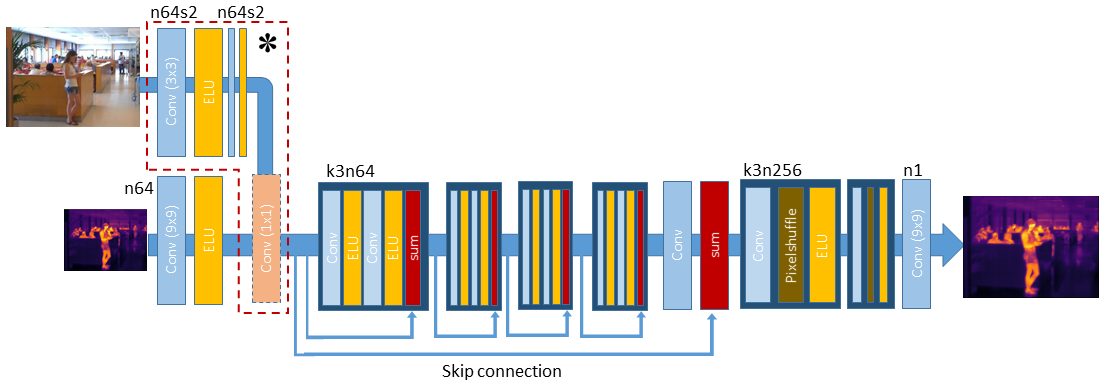}
        \caption{Generator}
    \end{subfigure}\\
    
    \begin{subfigure}[b]{\textwidth}
    \centering
        \includegraphics[width=.9\textwidth]{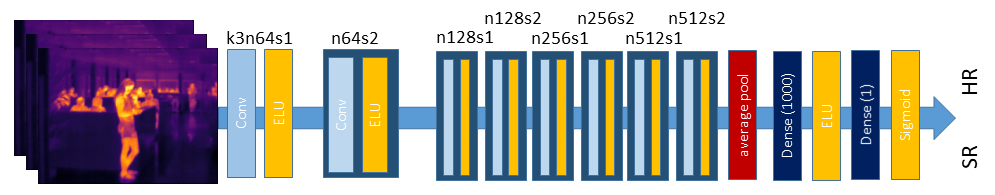}
        \caption{Discriminator}
    \end{subfigure}\quad
    
    \caption{Architecture of our Generator and Discriminator networks with the corresponding (k) kernel size, (n) number of channels and (s) their stride when it is changed. The highlighted area with (*) indicates the model that merges RGB and thermal channels}\label{fig:Model}
\end{figure}

\subsection{Network Architecture}
The generator baseline network thermal SRCNN (TSRCNN) shown in Fig.~\ref{fig:Model} consists of 5 identical residual blocks, inspired by ~\cite{2016arXiv160904802L} and follows the layout proposed by ~\cite{torch}. Each block consists of two convolutional layers with $3 \times 3$ kernel size, and 64 feature maps each followed by an ELU activation function ~\cite{clevert2015fast}. Batch-normalization layers ~\cite{ioffe2015batch} were removed from the residual block, since, as indicated in ~\cite{lim2017enhanced,huang2018densely}, they are unnecessary for the SR problem. It is also stated that once batch-normalization layers are used, they harm the generated output. To preserve the feature maps size, reflective padding is used around the boundaries before each convolution layer. Feature maps resolution is then increased using two sub-pixel convolution layers ~\cite{shi2016real}.

Visual RGB images are integrated into the model using two convolution layers with $3 \times 3$ kernel size and 64 feature maps each followed by an ELU activation function. Each convolution layer used a 2-step stride, which reduced the feature maps size to be the same size as the thermal input, before they are fused to form the visual-thermal SRCNN (VTSRCNN) model as shown in fig.~\ref{fig:Model} (*). The fusion is handled by concatenating the two feature maps, followed by a convolution layer with $1 \times 1$ to reduce channel dimensionality to 64 channels. Due to the high correlation between the visual-thermal modes and the rich texture information in the visual image, the network is supposed to learn these features and fuse them to produce SR thermal images with high perceptual quality. Giving an input thermal image $T \in \mathbb{R}^{H\times W}$ and an input RGB image $X \in \mathbb{R}^{H \times W \times 3}$, the objective function $\hat{Y} = F(T,X)$ seeks to map the LR thermal image to the HR thermal image $Y \in \mathbb{R}^{H\times W}$. 

The last two models are re-trained on the GAN based model to form the GAN proposed models (TSRGAN and VTSRGAN). To do this a discriminator network as shown in Fig \ref{fig:Model} is added, to classify generated images from original images. The network architecture is similar to the work in ~\cite{2016arXiv160904802L}, except for the batch normalization layer and ELU activation function. The model consists of eight convolution layers with $3 \times 3$ that increase the feature by a factor of 2 from 64 to 512 channels. Image resolution is reduced using a 2-step stride convolution between each layer which doubles the channels number. In this model, adapted average pooling is used on top of the 512 feature maps followed by two dense layers. Finally, a sigmoid activation function is used to produce a probability of the input being an original HR image or a generated SR thermal image.

\subsection{Loss Function}
Our baseline models (TSRCNN and VTSRCNN) are trained using only the content loss function, which is the mean square error (MSE), while our GAN based models (TSRGAN and VTSRGAN) are trained on a weighted sum of the content loss and the adversarial loss, which is obtained from the discriminator network. By using only the adversarial loss, the models are not able to converge. This is most likely ascribed to the lack of overlap in the distribution supports between original images and generated images. Therefore, the content loss was necessary for the GAN based model. The models that take only thermal images or visual-thermal images are shown in Eq.(\ref{eq:L1}).

\begin{equation}
T^{SR} = \begin{cases}
G(T^{LR}), &\text{only thermal model}\\
G(T^{LR}, RGB), &\text{visual-thermal model}
\end{cases}
\label{eq:L1}
\end{equation}

\textbf{Content Loss.} MSE in Eq.(\ref{eq:L2}) is the most used optimization function in the SR image problem \cite{2015arXiv150100092D,2016arXiv161109969Y,kim2016accurate}. The model is trained on optimizing the Euclidean distance between the constructed $T^{SR}$ and the ground truth $ T^{HR}$. Although the MSE is highly correlated in maximizing the PSNR, it suffers from the lack of high-frequency details, which results in blurred and over-smoothed images. However, it does help the model preserve low-frequency details from the LR image and supports the adversarial loss which could not always ensure convergence.

\begin{equation}
l_{MSE} =  \frac{1}{2MN}  \sum_{x=1}^{M} \sum_{y=1}^{N}(T^{HR}_{x,y} - T^{SR}_{x,y})^2 
\label{eq:L2}
\end{equation}

\textbf{Adversarial Loss.} To ensure high-frequency details from the original HR distribution, the adversarial loss is added to the content loss. The models are first pre-trained on the content loss MSE and then fine-tuned using the total loss function $l^{SR}$ (Eq.~\ref{eq:L3}), which is a weighted sum of the adversarial loss $ l_{gen}$ (Eq.~\ref{eq:L5}) and the content loss $ l_{MSE}$ (Eq.~\ref{eq:L1})where $\lambda = 1e-2$ is a fixed parameter.

\begin{equation}
l^{SR} = l_{gen} +  \lambda l_{MSE} 
  \label{eq:L3}
\end{equation}

The discriminator network is trained using the cross-entropy loss function (Eq.~\ref{eq:L4}) that classifies the original images from the generated images. The generator loss $ l_{gen}$ is trained on the recommended equation (Eq.13 in \cite{goodfellow2016nips}).

\begin{equation}
    J^{(D)} = E_{(T^{HR})}  log(D(T^{HR})) + E_{(T^{LR})}  log(1 - D(T^{SR})
    \label{eq:L4}
\end{equation}

\begin{equation}
  l_{gen} = J^{(G)} = - \frac{1}{2} E_{(T^{LR})}  log(D(T^{SR})
  \label{eq:L5}
\end{equation}

\subsection{Implementation and training details}
All of the models are implemented in Pytorch and trained on NVIDIA TITAN Xp using randomly selected mini-batches of size 12, plus 12 RGB mini-batches when the visual-thermal fusion model is used. The generator model uses RMSPROP optimizer with alpha=0.9. In the GAN based model the discriminator is trained using the SGD optimizer. The baseline model, and all other investigated models, are trained using the content loss for 5000 epochs. The pre-trained baseline model is used to initialize the generator in the adversarial model, where D and G are trained for another 5000 epochs. All models are trained with initial learning rate $10^{-4}$ and decreased to $10^{-6}$.


\section{Experiments}
\subsection{Model analysis}

\subsubsection{Resolution training size.}
\begin{figure}
	\centering
	\includegraphics[width=1\textwidth]{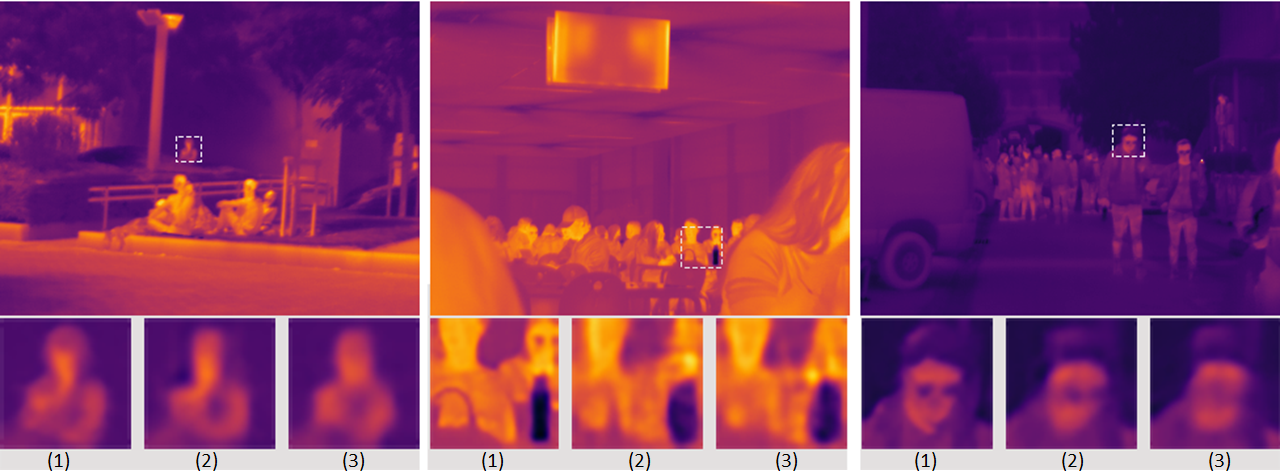}
	\caption{(1) HR image. (2) Our proposed model TSRCNN trained on LR image features with PSNR/SSIM (52.353/0.9495). (3) The same model trained on the up-sampled features using 2 Deconv layers with PSNR/SSIM (52.656/0.9510)}
	\label{fig:hr}
\end{figure}

Attention here is focused on showing the effect of training the model on LR features or on their up-sampled version, which is the difference between the two network schemes (c) and (d) shown in Fig \ref{fig:modelAbstract}. The first extracts and optimizes features of the original LR image, and up-samples them at the end of the network, while the second up-samples the input features first and then optimizes them along the network. By looking at the trade-off between the computation cost and the model performance, the trained model on the up-sampled features increased the computation cost and did not add a significant improvement to the generated images. Instead, the up-sampled training as shown in Fig.~\ref{fig:hr}, depicts a slight increment in the PSNR/SSIM values compared to our proposed model (TSRCNN) but the model could not generate some fine texture details such as the handbag handle in the second image and the person in the background of the first and third images.

\subsubsection{Evaluation with the state-of-the-arts.}
\begin{figure}
	\centering
	\includegraphics[width=1\textwidth]{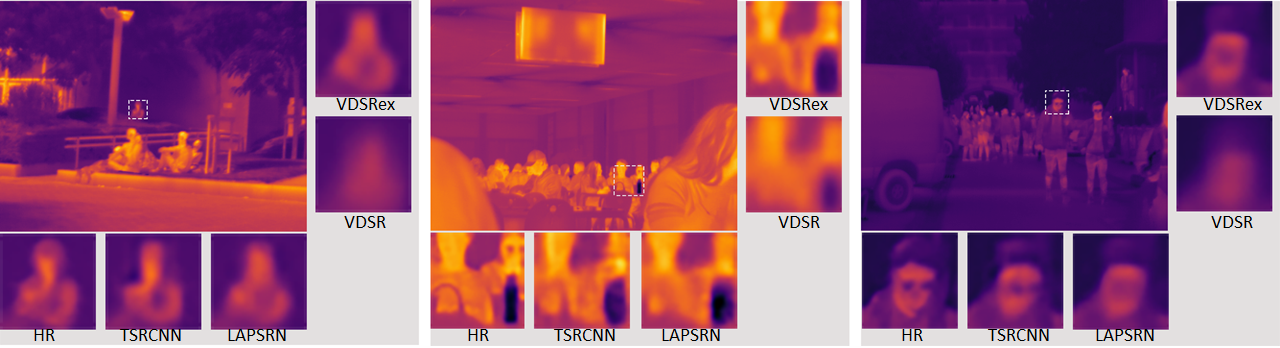}
	\caption{Comparison between our baseline model TSRCNN and state-of-the-art, LAPSRN \cite{lai2017deep}, VDSR ~\cite{kim2016accurate} and our extended VDSRex.}
	\label{fig:STOA}
\end{figure}

Before validating the residual learning model scheme and the up-sampling methods of the proposed models, the proposed baseline model is compared with state-of-the-art: VDSR \cite{kim2016accurate} and LAPSRN ~\cite{lai2017deep} which are based on residual learning. VDSR is implemented in two models, (1) the original VDSR that takes only thermal images as input, and (2) our extended VDSRex that takes visual-thermal images as an input of 4 channels. LAPSRN is trained using only thermal images. The experiment was run on our ULB17-VT benchmark, using the same size model and training procedure explained in the STOA paper. Fig.~\ref{fig:STOA} shows that VDSR failed to produce high-frequency details, while the VDSRex produced better results taking advantage of the visual-thermal fusion. However, the proposed baseline model generates images with sharp details and higher perceptual image quality. Table.~\ref{table:PSNR} shows that the proposed model also obtained higher PSNR/SSIM value than the STOA.

\subsubsection{Residual learning and Up-sampling methods.}
Attention was then focused on adapting and investigating the residual learning model to the thermal SR problem. To use the baseline model, TSRCNN in this model scheme, the input should be rescaled to have the same size as the residual output. We trained four different models with four different up-sampling methods: (1) InpDconv-TRSCNNres that integrates TRSCNN and two deconvolution layers to up-sample the input image; (2) InpBilin-TRSCNNres that up-samples the input using bilinear interpolation; (3) InpBicub-TRSCNNres that uses bicubic interpolation; (4) AllDconv-TRSCNNres which is similar to (1) but the two Pixelshuffle layers at the end of the network are replaced by two deconvolution layers.

\begin{figure}
	\centering
	\includegraphics[width=1\textwidth]{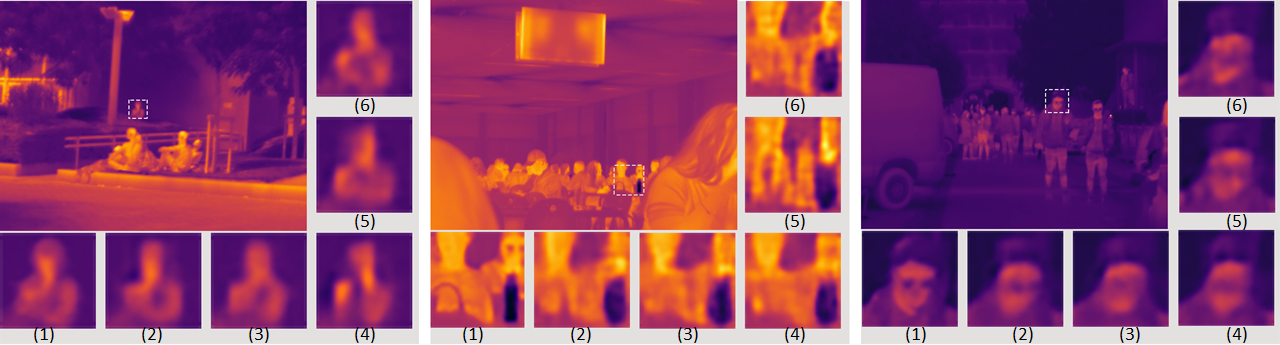}
	\caption{(1) HR image. (2) Our proposed TSRCNN model with no residual learning and (52.353/0.9495). (3) InpDconv-TSRCNNres with (52.323/0.9491). (4) AllDconv-TSRSCNNres with (52.430/0.9495). (5) InpBilin-TSRCNNres with (51.023/0.9415). (6) InpBicub-TSRCNNres with (51.670/0.9442).  Values between brackets are PSNR/SSIM.}
	\label{fig:ups}
\end{figure}

Fig.~\ref{fig:ups} shows that the models trained on bilinear and bicubic interpolations methods failed to produce comparable perceptual quality results, and have the lowest PSNR/SSIM values. Models (1) and (4) that use trainable up-sampling methods produced better perceptual results with high PSNR/SSIM values. However, the proposed model produced sharper edges and finer details. Note that the person in the first and third images has sharper details in the proposed model, the bag handle exists only in our model.
To this end and to better evaluate the contribution of the models, all models are taken into our qualitative evaluation study.  

\subsubsection{Visual-Thermal fusion.}
To demonstrate the effect of integrating visual-thermal fusion in the SR problem, and to investigate if the rich information in the visual images can help to produce better thermal SR images, the baseline thermal SR Convnet model (TSRCNN) was trained on only thermal images using the network architecture shown in Fig.~\ref{fig:Model} (a).Also, the model Visual-Thermal SR Convnet (VTSRCNN)model was trained on thermal and visual images using the model shown in Fig.~\ref{fig:Model} (a) which integrate the branch (*). Fusing visual-thermal images added more details to the produced thermal SR images, but also added some artifacts in parts of the images. In particular, these artifacts appear when there is displacement in the objects due to the camera design and image capturing mechanism. The integration enhanced the SR images slightly. Thus the comparison is difficult between them, but it can be seen as sharp and extra details in small regions of the SR images as shown in Fig.~\ref{fig:opt}. Therefore, a qualitative evaluation study was set to validate the contribution of the visual-thermal fusion. 

\subsubsection{Optimization function.}
\begin{figure}
	\centering
	\includegraphics[width=1\textwidth]{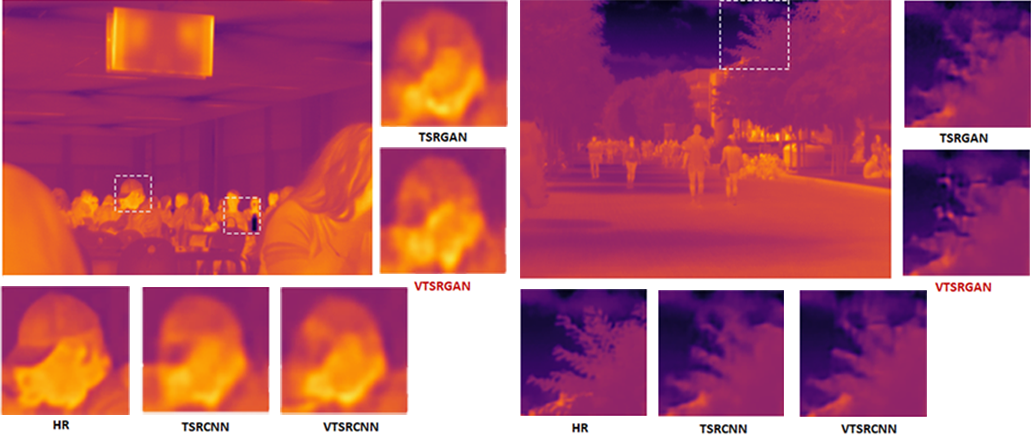}
	\caption{Our proposed models trained only on thermal or on visual-thermal fusion, using only content loss or with adversarial loss.}
	\label{fig:opt}
\end{figure}

To investigate the contribution of the adversarial loss on producing high perceptual quality SR images; the two models (TSRGAN that takes only thermal images and VTSRGAN which takes visual-thermal images) are trained using content loss and adversarial loss. Fig.~\ref{fig:opt} shows the proposed models and their contribution to enhancing the thermal SR perceptual quality. Models trained with adversarial loss produced images with high texture details and high-frequency information. Although they added some small artifacts, they produced images that are sharper and less blurry than images generated using only content loss. Table.~\ref{table:PSNR} shows the relationship between the mean square error and the PSNR validation measurement, the model TSRCNN has the highest PSNR/SSIM value and also the most blurry images. To better validate the perceptual quality, our four models are added to the qualitative evaluation study.

\begin{table}
\caption{Quantitative evaluation of the proposed models and the STOA}
\label{table:PSNR}
\begin{tabular}{|l|l|l|l|l|}
\hline
Our models $\qquad\quad$& TSRCNN & VTSRCNN & TSRGAN & VTSRGAN  \\
\hline
PSNR/SSIM  & 52.353/0.9495 & 51.727/0.9434 & 51.094/0.9285 & 50.979/0.9289 \\
\hline
STOA $\qquad\qquad$ & VDSR & VDSRex & LAPSRN & \\
\hline
PSNR/SSIM  & 45.557/0.8328 & 52.027/0.9395 &  51.936/0.9526 & \\
\hline
\end{tabular}
\end{table}

\subsection{Qualitative evaluation study}
To evaluate the proposed models and the different investigated schemes in comparison with the STOA, a qualitative evaluation study to alleviate the PSNR/SSIM impact and to assist with evaluating the human visual perception was conducted. A website that allows users to choose the most similar image to its original HR counterpart was created. Twenty-two people, with and without computer vision backgrounds, contributed to this evaluation process. They were asked to vote for a large study case, the test set used in the ULB17-VT benchmark with 58 samples. For each image, 9 models were selected for this evaluation process.

Running a qualitative evaluation study on 9 models with 58 images for each is very exhaustive work for the raters. To encourage them and to reduce the overall number of the selections required, three evaluation groups were created. In each group and for each image only three models were presented, these models were selected randomly and not repeated. The evaluation page shows the original HR image and the three selected models output. The user was asked to select the image that was most similar to its original HR image counterpart. For each selection process, a +1 was awarded in favor of the chosen model against the two other models shown. For example, in group 1 image 1 the models [4, 5, 6] outputs are presented. If the user selects the second image, this means the ranking output is +1 for model 5 over model 4 $(f_{54}) = +1$ and +1 for model 5 over model 6 $(f_{56}) = +1$. Finally, the total votes for and against the paired models are normalized as shown in Eq.(\ref{eq:L6}), and also normalized by the number of times these paired models were presented in the evaluation process.

\begin{align}
f_{ij} &= \begin{cases}
    i-j,& \text{if } f_{ij} > f_{ji}\\
    0,              & \text{otherwise}  
\end{cases}  & \text{} 
f_{ji} &= \begin{cases}
    j-i,& \text{if } f_{ji} > f_{ij}\\
    0,              & \text{otherwise}  
\end{cases} & \text{}
 \label{eq:L6}
\end{align}

\begin{figure}
	\centering
	\includegraphics[width=.9\textwidth]{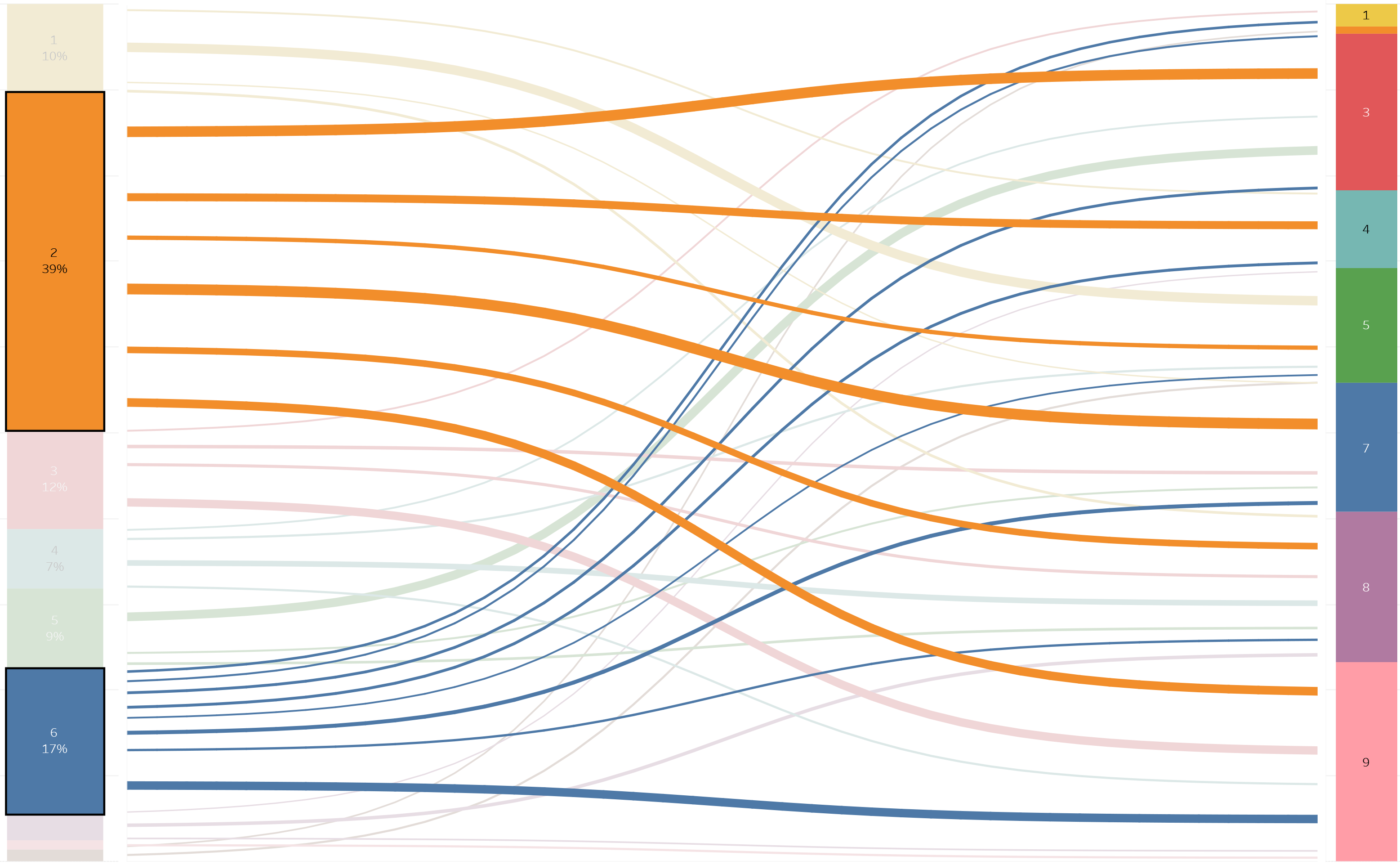}
	\caption{Color-coded votes flow diagram of models: (1)TSRCNN, (2)VTSRCNN, (3)InpDconv-TRSCNNres, (4)AllDconv-TRSCNNres, (5)TSRGAN, (6)VTSRGAN, (7)VDSR, (8)VDSRex, (9)LAPSRN. Left/right represent the models. Paths in the middle represent the vote in favor between model.}
	\label{fig:Sankey}
\end{figure}

The color-coded votes diagram shown in Fig~.\ref{fig:Sankey} shows that the proposed models, which integrate visual-thermal fusion, are the highest selected models against almost all the other models. The size of the models to the left of the graph indicates the number of times these models were voted in favor over all the other models, it shows that the model VTSRCNN has $39\%$ and VTSRGAN has $17\%$. The larger the model to the right of the graph indicates the number of times these models were voted against. The weight of the paths indicates the number of times these models were selected in favor against the opposite model. Our human visual perception study shows that the proposed models with visual-thermal fusion have the highest votes in favor and the lowest votes in disfavor. This highlights the benefits of integrating visual-thermal fusion in the thermal super-resolution problem.

\subsection{Limitations}
Although the proposed models generate better thermal SR images in human visual perception, some artifacts were noticed in the generated images. These artifacts are most likely caused by the device design or are due to the displacement of the device or the object. The models can preserve the high-frequency details of the visual images, the displacement problem could not be approached simply, and needs a better synchronized device to overcome the problem. Due to this displacement in some samples, the reconstructed versions suffer from artifacts around these objects compared to images with no displacement. We leave this problem open for further study and investigation.


\section{Conclusion}
In this paper, the problem of thermal super-resolution enhancement using the domain of visual images was addressed.  A deep residual network that provides a better solution compared to other network schemes and training methods in the literature was proposed. Our result highlights that visual-thermal fusion can enhance the thermal SR image quality, also by the contribution of GAN based model. Furthermore, a qualitative evaluation study was performed and analyzed. This evaluation  indicates a better understanding of the problem evaluation than the widely used PSNR/SSIM measurements. Lastly, a new visual-thermal benchmark in super-resolution problem domain was set.

%
%
\bibliographystyle{splncs04}
\bibliography{egbib}

\begin{thebibliography}{10}
\providecommand{\url}[1]{\texttt{#1}}
\providecommand{\urlprefix}{URL }
\providecommand{\doi}[1]{https://doi.org/#1}

\bibitem{chen2016color}
Chen, X., Zhai, G., Wang, J., Hu, C., Chen, Y.: Color guided thermal image
  super resolution. In: Visual Communications and Image Processing (VCIP),
  2016. pp.~1--4. IEEE (2016)

\bibitem{chen2017face}
Chen, Z., Tong, Y.: Face super-resolution through wasserstein gans. arXiv
  preprint arXiv:1705.02438  (2017)

\bibitem{cho2014multi}
Cho, H., Seo, Y.W., Kumar, B.V., Rajkumar, R.R.: A multi-sensor fusion system
  for moving object detection and tracking in urban driving environments. In:
  Robotics and Automation (ICRA), 2014 IEEE International Conference on. pp.
  1836--1843. IEEE (2014)

\bibitem{choi2016thermal}
Choi, Y., Kim, N., Hwang, S., Kweon, I.S.: Thermal image enhancement using
  convolutional neural network. In: Intelligent Robots and Systems (IROS), 2016
  IEEE/RSJ International Conference on. pp. 223--230. IEEE (2016)

\bibitem{clevert2015fast}
Clevert, D.A., Unterthiner, T., Hochreiter, S.: Fast and accurate deep network
  learning by exponential linear units (elus). arXiv preprint arXiv:1511.07289
  (2015)

\bibitem{2015arXiv150100092D}
{Dong}, C., {Change Loy}, C., {He}, K., {Tang}, X.: {Image Super-Resolution
  Using Deep Convolutional Networks}. ArXiv e-prints  (Dec 2015)

\bibitem{goodfellow2016nips}
Goodfellow, I.: Nips 2016 tutorial: Generative adversarial networks. arXiv
  preprint arXiv:1701.00160  (2016)

\bibitem{goodfellow2014generative}
Goodfellow, I., Pouget-Abadie, J., Mirza, M., Xu, B., Warde-Farley, D., Ozair,
  S., Courville, A., Bengio, Y.: Generative adversarial nets. In: Advances in
  neural information processing systems. pp. 2672--2680 (2014)

\bibitem{torch}
Gross, S., Wilber, M.: Training and investigating residual nets.
  https://torch.ch/blog/2016/02/04/resnets.html (2016)

\bibitem{huang2018densely}
Huang, Y., Qin, M.: Densely connected high order residual network for single
  frame image super resolution. arXiv preprint arXiv:1804.05902  (2018)

\bibitem{hwang2015multispectral}
Hwang, S., Park, J., Kim, N., Choi, Y., Kweon, I.S.: Multispectral pedestrian
  detection: Benchmark dataset and baselines. In: Proceedings of IEEE
  Conference on Computer Vision and Pattern Recognition (CVPR) (2015)

\bibitem{ioffe2015batch}
Ioffe, S., Szegedy, C.: Batch normalization: Accelerating deep network training
  by reducing internal covariate shift. arXiv preprint arXiv:1502.03167  (2015)

\bibitem{isola2017image}
Isola, P., Zhu, J.Y., Zhou, T., Efros, A.A.: Image-to-image translation with
  conditional adversarial networks. arXiv preprint  (2017)

\bibitem{johnson2016perceptual}
Johnson, J., Alahi, A., Fei-Fei, L.: Perceptual losses for real-time style
  transfer and super-resolution. In: European Conference on Computer Vision.
  pp. 694--711. Springer (2016)

\bibitem{kim2016accurate}
Kim, J., Kwon~Lee, J., Mu~Lee, K.: Accurate image super-resolution using very
  deep convolutional networks. In: Proceedings of the IEEE Conference on
  Computer Vision and Pattern Recognition. pp. 1646--1654 (2016)

\bibitem{kim2016deeply}
Kim, J., Kwon~Lee, J., Mu~Lee, K.: Deeply-recursive convolutional network for
  image super-resolution. In: Proceedings of the IEEE conference on computer
  vision and pattern recognition. pp. 1637--1645 (2016)

\bibitem{kiran2017single}
Kiran, Y., Shrinidhi, V., Hans, W.J., Venkateswaran, N.: A single-image
  super-resolution algorithm for infrared thermal images. INTERNATIONAL JOURNAL
  OF COMPUTER SCIENCE AND NETWORK SECURITY  \textbf{17}(10),  256--261 (2017)

\bibitem{lai2017deep}
Lai, W.S., Huang, J.B., Ahuja, N., Yang, M.H.: Deep laplacian pyramid networks
  for fast and accurate super-resolution. In: Proc. IEEE Conf. Comput. Vis.
  Pattern Recognit. pp. 624--632 (2017)

\bibitem{2016arXiv160904802L}
{Ledig}, C., {Theis}, L., {Huszar}, F., {Caballero}, J., {Cunningham}, A.,
  {Acosta}, A., {Aitken}, A., {Tejani}, A., {Totz}, J., {Wang}, Z., {Shi}, W.:
  {Photo-Realistic Single Image Super-Resolution Using a Generative Adversarial
  Network}. ArXiv e-prints  (Sep 2016)

\bibitem{ledig2016photo}
Ledig, C., Theis, L., Husz{\'a}r, F., Caballero, J., Cunningham, A., Acosta,
  A., Aitken, A., Tejani, A., Totz, J., Wang, Z., et~al.: Photo-realistic
  single image super-resolution using a generative adversarial network. arXiv
  preprint  (2016)

\bibitem{lim2017enhanced}
Lim, B., Son, S., Kim, H., Nah, S., Lee, K.M.: Enhanced deep residual networks
  for single image super-resolution. In: The IEEE conference on computer vision
  and pattern recognition (CVPR) workshops. vol.~1, p.~4 (2017)

\bibitem{panagiotopoulou2008super}
Panagiotopoulou, A., Anastassopoulos, V.: Super-resolution reconstruction of
  thermal infrared images. In: Proceedings of the 4th WSEAS International
  Conference on REMOTE SENSING (2008)

\bibitem{qu2017active}
Qu, Y., Zhang, G., Zou, Z., Liu, Z., Mao, J.: Active multimodal sensor system
  for target recognition and tracking. Sensors  \textbf{17}(7), ~1518 (2017)

\bibitem{shi2016real}
Shi, W., Caballero, J., Husz{\'a}r, F., Totz, J., Aitken, A.P., Bishop, R.,
  Rueckert, D., Wang, Z.: Real-time single image and video super-resolution
  using an efficient sub-pixel convolutional neural network. In: Proceedings of
  the IEEE Conference on Computer Vision and Pattern Recognition. pp.
  1874--1883 (2016)

\bibitem{wu2017srpgan}
Wu, B., Duan, H., Liu, Z., Sun, G.: Srpgan: Perceptual generative adversarial
  network for single image super resolution. arXiv preprint arXiv:1712.05927
  (2017)

\bibitem{2016arXiv161109969Y}
{Yang}, C., {Lu}, X., {Lin}, Z., {Shechtman}, E., {Wang}, O., {Li}, H.:
  {High-Resolution Image Inpainting using Multi-Scale Neural Patch Synthesis}.
  ArXiv e-prints  (Nov 2016)

\bibitem{zhang2017stackgan}
Zhang, H., Xu, T., Li, H., Zhang, S., Huang, X., Wang, X., Metaxas, D.:
  Stackgan: Text to photo-realistic image synthesis with stacked generative
  adversarial networks. In: IEEE Int. Conf. Comput. Vision (ICCV). pp.
  5907--5915 (2017)

\bibitem{zhu2017unpaired}
Zhu, J.Y., Park, T., Isola, P., Efros, A.A.: Unpaired image-to-image
  translation using cycle-consistent adversarial networks. arXiv preprint
  arXiv:1703.10593  (2017)

\end{thebibliography}

\end{document}